\documentclass[10pt,twocolumn,letterpaper]{article}

\usepackage{wacv}              

\usepackage{graphicx}
\usepackage{amsmath}
\usepackage{amssymb}
\usepackage{booktabs}
\usepackage{multirow}
\usepackage[dvipsnames]{xcolor}

\usepackage[pagebackref,breaklinks,colorlinks]{hyperref}

\usepackage[capitalize]{cleveref}
\crefname{section}{Sec.}{Secs.}
\Crefname{section}{Section}{Sections}
\Crefname{table}{Table}{Tables}
\crefname{table}{Tab.}{Tabs.}

\def\wacvPaperID{1904} 
\def\confName{WACV}
\def\confYear{2025}

\usepackage{filecontents}

\begin{document}

\title{Clothes-Changing Person Re-identification Based On Skeleton Dynamics}

\author{Asaf Joseph\ \ \ \ \ \  \ \  Shmuel Peleg\\
The Hebrew University of Jerusalem\\
}
\maketitle

\begin{abstract}

Clothes-Changing Person Re-Identification (ReID) aims to recognize the same individual across different videos captured at various times and locations. This task is particularly challenging due to changes in appearance, such as clothing, hairstyle, and accessories. We propose a Clothes-Changing ReID method that uses only skeleton data and does not use appearance features. Traditional ReID methods often depend on appearance features, leading to decreased accuracy when clothing changes. Our approach utilizes a spatio-temporal Graph Convolution Network (GCN) encoder to generate a skeleton-based descriptor for each individual. During testing, we improve accuracy by aggregating predictions from multiple segments of a video clip. Evaluated on the CCVID dataset with several different pose estimation models, our method achieves state-of-the-art performance, offering a robust and efficient solution for Clothes-Changing ReID.
\end{abstract}

\section{Introduction}
\label{sec:intro}

Person Re-Identification (ReID) aims to match individuals appearing in different videos, taken at different times and locations. Two common ReID cases are (i) Same-Clothes and (ii) Clothes-Changing. In Same-Clothes ReID, a person keeps the same appearance, including clothing, hairstyle, and accessories. Several methods give excellent results in this setup \cite{hermans2017defense, xiao2017margin, 6909421}.
These methods rely on appearance features, and perform poorly in the Clothes-Changing setup.

\begin{figure}[tbh]
  \centering
   \includegraphics[width=\linewidth]{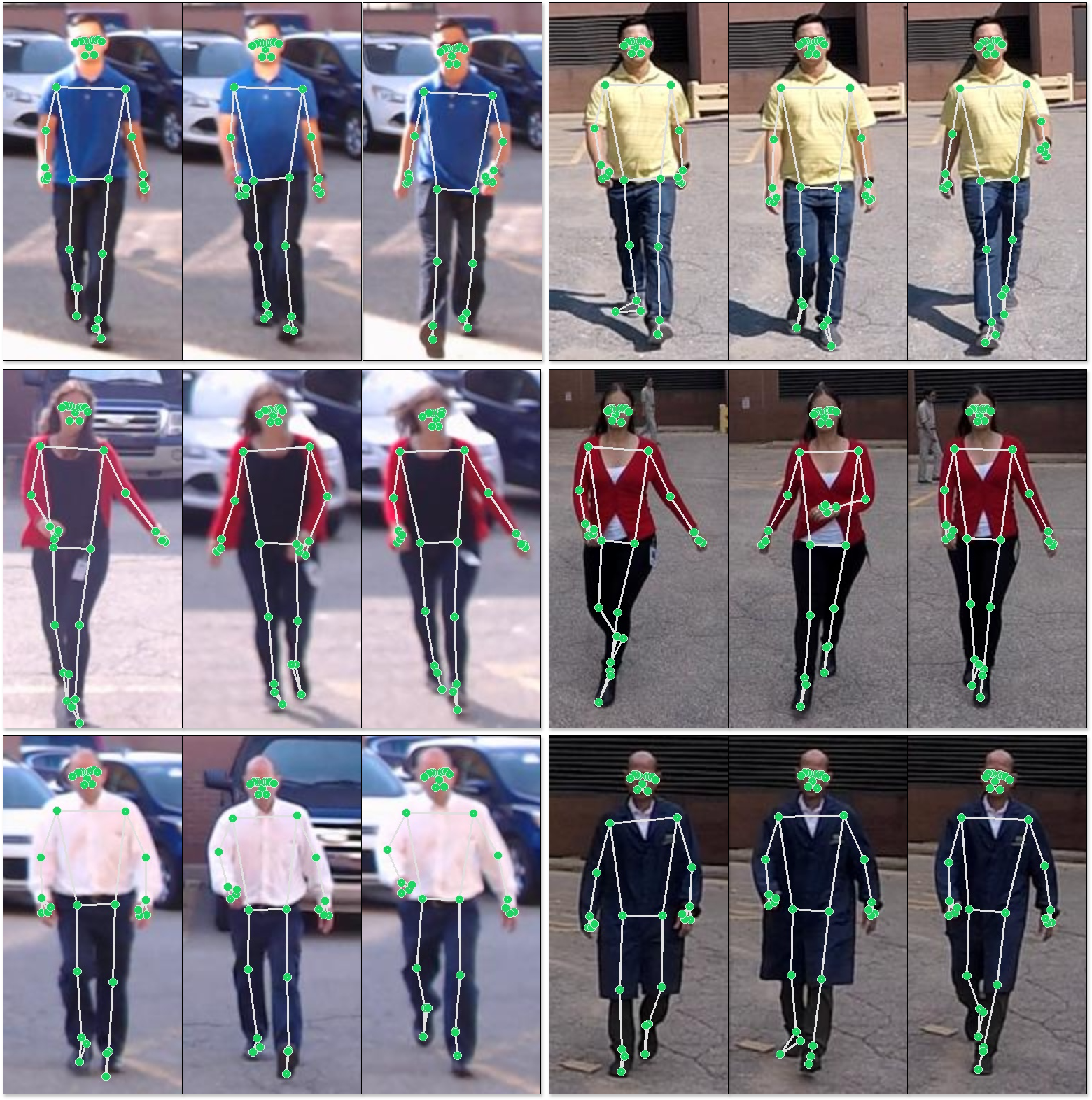}
   \caption{The information in Skeletons: Each row shows the same person, where the three left and three right video frames were taken months apart. Despite changes in clothing and appearance, the hands movements remains consistent and identifiable. This consistent skeletal motion forms the basis of our method, enabling robust person re-identification despite changes in appearance.}
   \label{fig:same-person-different-interval}
\end{figure}

\begin{figure*}[tbh]
\centering
\includegraphics[width=0.68\linewidth]{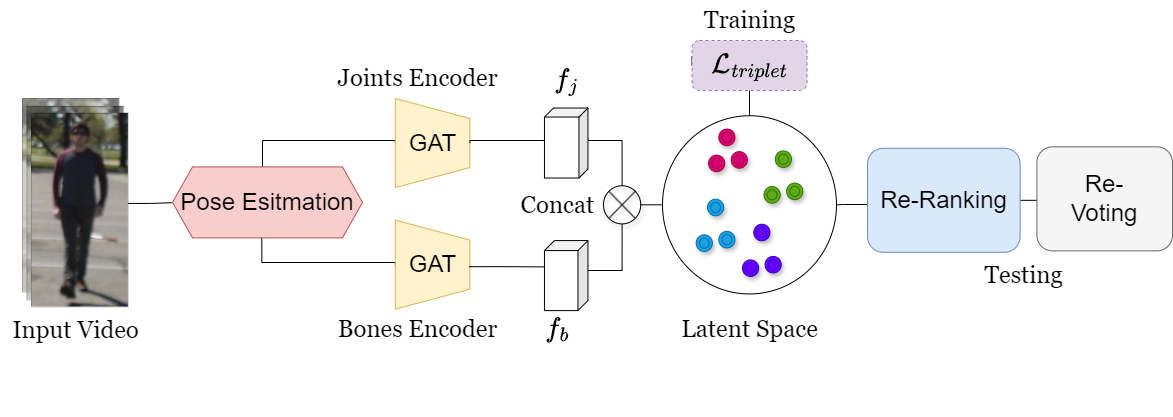}
\caption{
Method Overview: An input video with $T$ frames is first processed by a pose estimation model, then fed into two streams of Graph Convolution Networks (GCNs) to extract features $f_j, f_b$ for the joints and bones streams, respectively. These features encode the video into a shared latent space. During training, the model is optimized using the triplet loss. During testing, two additional statistical methods, Re-Ranking and Re-Voting, are applied to extract the final ranked matching.}
\label{fig:method}
\end{figure*}

Some methods have attempted to learn clothing-invariant features, such as silhouettes \cite{9577677, jin2022cloth}, contour sketches \cite{Yang_2021}, or focusing on fine-grained features such as individual body parts \cite{sun2018beyond, ni2023part}. Other approaches have focused on extracting temporal information from video datasets to learn additional time-dependent features and integrating these with spatial features \cite{9156954}.

In \cite{fan2020learning}, the authors introduced the problem of leaking personal information in ReID models. They showed how models that rely on appearance features are exposed to personal information such as skin color and environmental clues about an individual's work and living space. To overcome this drawback, the authors suggested performing ReID in the skeleton space, where the only information used is how an individual's body is shaped and moves.

The privacy concerns, and the advancements in the Skeleton-Based Action-Recognition task \cite{duan2022revisiting}, inspired  methods to use the human body pose also for ReID \cite{10483285, qian2020long}.
Action recognition is orthogonal to ReID in the sense that Action Recognition should be invariant to the identity of the individual, while ReID should be invariant to the individual's activity. Although these methods show promising results, all of them still add appearance features in addition to the skeleton.

More recently, an innovative paper \cite{Nguyen_2024_CVPR} suggested
data augmentation to enrich the dataset with various clothing and poses. \cite{Nguyen_2024_WACV} innovated the use of 3D mesh of the human body.
Results presented in these two papers are promising, but they also use image appearance data, conflicting with privacy concerns.


In this work, we propose a Graph Convolution Network (GCN) to learn spatial (body shape) and temporal (body motion) information based only on skeleton information. Using only the skeleton data reduces dimensionality and satisfies privacy concerns. We show that using statistical methods such as Re-Ranking \cite{zhong2017re} and Ranked-Voting during test time can significantly improve the model performance. We obtain state-of-the-art results on the CCVID dataset, one of the most famous, publicly available datasets for Clothes-Changing ReID.

\section{Related Work}
\label{sec:related-work}

\subsection{Person Re-Identification (Re-ID)}

Early ReID papers addressed primarily the Same-Clothes setup, where an individual's appearance does not change much. Initially, handcrafted features were designed to capture the unique characteristics of individuals, including elements such as color and texture \cite{5539926, Liao_2015_CVPR}. Handcrafted features were later replaced by trained features using deep models \cite{Ahmed_2015_CVPR, 6909421}.

\subsection{Clothes-Changing Re-ID (CC-ReID)}

The Same-Clothes ReID setup does not hold in many real-world applications, where ReID is needed after periods longer than one day. In these cases cloths, accessories, hair style, and more are likely to change. To overcome these temporal changes, features that are invariant to changes in an individual's appearance are required. An additional challenge in Clothes-Changing ReID is the lack of publicly available datasets for developing new methods. This led the focus towards video-based datasets \cite{xu2023deepchange, davila2023mevid, gu2022clothes}, where the temporal information can also be taken into account. 
3D-CNN's \cite{liao2019video, gu2020appearance, li2019multi}, RNN's \cite{yan2016person, xu2017jointly}, and GCN's \cite{Wu_2020, 9156954, Nguyen_2024_WACV} were used to learn clothes-invariant spatial-temporal features. In \cite{Nguyen_2024_CVPR}, the authors suggested an auxiliary task of 3D human mesh prediction to enhance the appearance features. In \cite{arkushin2022reface}, face features were used as time-invariant features, in addition to the appearance features, to enhance the clothes-invariant properties.

Most of the above methods use some appearance features. In this paper, We also introduce simple statistical methods at test time to improve model performance. Since our model is exposed only to skeleton information, it reduces the amount of personal information learned by the model.

\section{Proposed Method}
\label{sec:proposed-method}

\subsection{Overview}

Video-Based Clothes-Changing ReID datasets are represented as $\mathcal{D}: \{\vec{X}, \vec{Y}\}_{i=1}^N$, where $\vec{X}$ is a video, 
$\vec{Y}$ is the person, camera and clothes ID's for the given video,
and $N$ is the total number of individuals.

An overview of our proposed method is shown in Fig \ref{fig:method}. A pose estimation process (e.g. 
BlazePose \cite{bazarevsky2020blazeposeondevicerealtimebody} or 
OpenPose \cite{cao2017realtime}) 
computes the skeleton of each person in the videos. A Graph-Convolution-Network (GCN) is trained to encode the sequence of $K$ skeletons into a latent space, mapping  a video segment of $K$ successive frames of the same individual into similar latent values, while video segments of other individuals are mapped into much different values. In this paper we used $K=50$.

During inference each query video is processed by a pose estimator, and the trained GCN is applied on the skeletons of each video segment of length $K$. The latent value of each video segment is matched to the segments in the gallery to find the best matches.
Following the initial matching, Re-Ranking and Re-Voting methods are used to improve the results by leveraging the multiple video segments for each video both in the query and in the gallery.

\subsection{Graph Representation of Skeleton Sequences}

\begin{figure}[tb]
  \centering
   \includegraphics[width=0.3\linewidth]{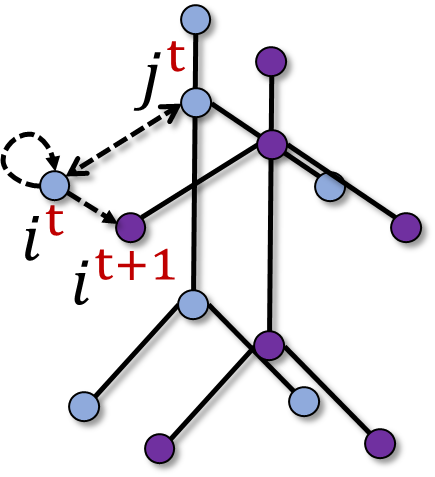}
   \caption{Schematic graph representation of skeleton sequences: Spatial edges connecting joints like \(i^t\) and \(j^t\) based on human body connectivity. Temporal edges linking all joints \(i\) between frames \(t\) and \(t+1\). Each node is also connected to itself.}
   \label{fig:skeleton-graph}
\end{figure}

A sequence of skeletons is represented by a spatio-temporal graph as shown schematically in Fig.~\ref{fig:skeleton-graph}. The connections in the spatial dimension are established based on the natural connectivity of the human body as provided by the BlazePose model \cite{bazarevsky2020blazeposeondevicerealtimebody}. We were inspired by the graph construction presented by the original AAGCN paper \cite{shi2019two}. 

There are two parallel graphs, a \textbf{joints graph} and a \textbf{bones graph}. A node $v_i$ in the joints graph is a vector $(x_i, y_i, z_i, c_i)$ that encodes the 3D location of the node and its confidence as given by the pose detection model. A node $v_k$ in the bones graph is defined by the two joints $v_i$ and $v_j$ it connects, $v_k : (x_j - x_i, y_j - y_i, z_j - z_i, max(c_j, c_i))$.

For each of the above graphs we have three $N \times N$ spatial adjacency matrices $\mathbf{A}_k$, $k=0,1,2$. $k=0$ indicate incoming links, $k=1$ indicate outgoing links, and $k=2$ indicate self loop. $\mathbf{A}_0$ is normalized such that all incoming links sum to 1, and  $\mathbf{A}_1$ is normalized such that all outgoing links sum to 1.

In the temporal dimension, each vertex $v_i^t$ at time $t$ is connected to its consecutive vertex $v_i^{t+1}$ at time $t+1$.

\begin{figure}[tb]
  \centering
   \includegraphics[width=0.9\linewidth]{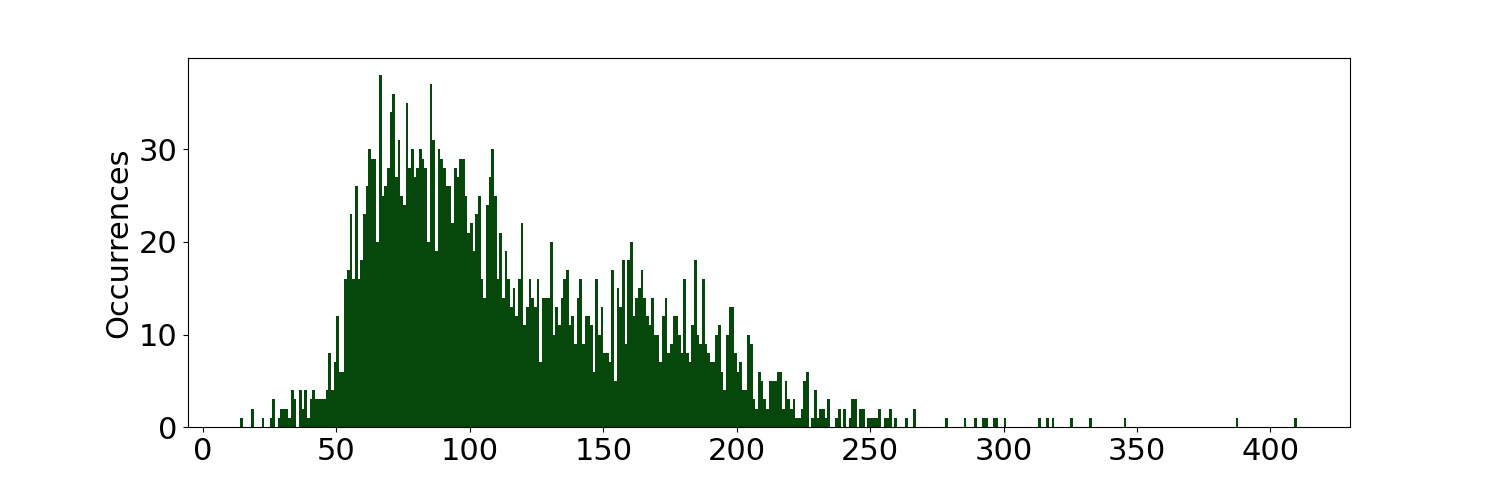}
   \caption{The distribution of the number of frames per video clip across the entire CCVID dataset. The X-axis represents the number of frames in each video clip, while the Y-axis shows the number of videos corresponding to each frame count.}
   \label{fig:ccvid-frames-distribution}
\end{figure}

The video clips in CCVID have different lengths, whose distribution is shown in 
Fig. \ref{fig:ccvid-frames-distribution}. Based on this distribution we have decided to divide the video into multiple overlapping segments of 50 frames each, with an overlap of 25 frames. Segments longer than 50 would exclude many shorter videos. Since 50 frames cover more than 3 seconds (at 16 frames per second), it is long enough to cover a couple of walking steps.

\subsection{Spatio-Temporal GCN Encoder}
\label{sec:GCN}

Our graph encoder $E$ consists of several layers based on a simplified version of the Adaptive Graph Convolutional Network (AAGCN) \cite{shi2019two} developed for skeleton-based action recognition. In this approach, one GCN is trained for the joints graph and another GCN is trained for the boned graphs.

Let the layer $\mathbf{f}^0$ represent the initial graph, where each node has 4 parameters $(x, y, z, c)$. The full operation of the GCN from Layer $l$ to Layer $(l+1)$ in the spatial dimension can be expressed as:

\begin{equation} 
    \mathbf{f}^{l+1} = \sum_{k=0}^{2} \mathbf{W^l}_k \times \mathbf{f}^{l} \times \mathbf{A}_k 
    \label{eq:gcn}
\end{equation} 

Where $\mathbf{f}^{l}$ represents the feature nodes at layer $l$, $\mathbf{W^l}_k$ corresponds to the learned weights transferring from layer $l$ to layer $(l+1)$, $\mathbf{A}_k$ are the three $N \times N$ adjacency matrices, and the $\times$ symbol denotes the matrix multiplication operation. The weights $\mathbf{W^l}_k$ increase the number of channels at each node from 4 channels at the input graph to 256 channels after 5 GCN layers, done in the following order:  [4, 16, 32, 64, 128, 256].

After building each GCN layer on each skeleton, a $9 \times 1$ temporal convolution is performed on all nodes, with stride of 2. Input level 0 has 50 skeletons with 4 channels per node, Level 1 has 25 skeletons and 16 channels, etc. The dimensions of the data at each node from level 0 to level 5 are [50$\times$4, 25$\times$16, 12$\times$32, 6$\times$64, 3$\times$128, 1$\times$256],

The last layer of the GCN has 256 channels (features) per each of the 33 nodes in the skeleton. We generate a descriptor of 256 features for each skeleton graph by computing for each feature the $L_3$ norm over all nodes, defined as:
\begin{equation}
    \| x \|_3 = \left( \sum_{i=1}^n |x_i|^3 \right)^{\frac{1}{3}}
    \label{eq:lp-pool}
\end{equation}
$L_3$ norm gave better results than other norms, e.g. average pooling or max pooling.

The descriptors from the joints graph and the bones graph are concatenated to generate a 512 long descriptor for each 50 frames segment.

In training the final descriptor is used to optimize the triplet-loss function, defined by:

\begin{equation}
    \mathcal{L}(a, p, n) = \max\left(0, d(a, p) - d(a, n) + \alpha\right)
    \label{eq:triplet-loss}
\end{equation}

Where $d(x, y)$ is the cosine distance function between $x$ and $y$, $a$, $p$, and $n$ are the anchor, positive, and negative examples, respectively. $\alpha$ is a margin parameter, we used $\alpha = 0.3$.

\subsection{Re-Ranking and Re-Voting}
\label{sec:rerank}

\begin{figure}[t]
  \centering
   \includegraphics[width=0.8\linewidth]{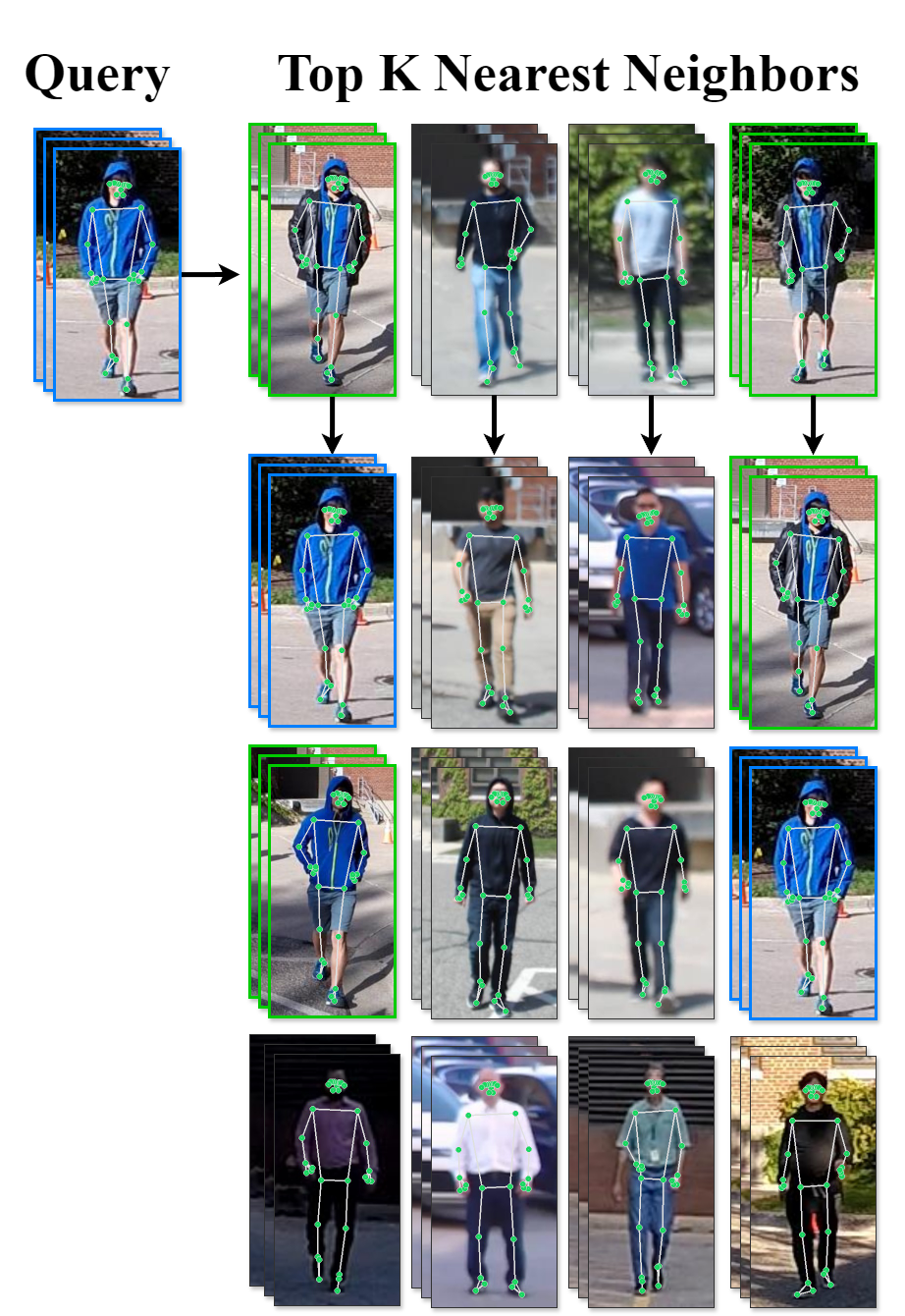}
   \caption{Re-Ranking Example: The top row shows the query image in blue frame, and its top 4 Nearest Neighbors (4-NN). Positive examples are in green frames, and negative examples are in black frames. Below each neighbor are its own 4-NN. We observe that the positive examples have the query as their own 4-NN. Consequently, K-NN neighbors that have the query as their own K-NN will be re-ranked higher than neighbors that do not have the query as their own K-NN.}
   \label{fig:reranking}
\end{figure}

\textbf{Re-Ranking}. In \cite{zhong2017re} an unsupervised re-ranking algorithm was proposed to improve the ReID task, based on $K$-reciprocal nearest neighbors (K-NN). In this algorithm, for each query image we find its $K$ nearest neighbors in the gallery, and for each initial K-NN neighbor we find its own $K$ nearest neighbors in the queries. An initial neighbor with the query as its own K-NN will be re-ranked higher than  An initial neighbor that does not have the query as one of its own K-NN. The more instances of the query in the reciprocal NN the better. This is similar to Best-Buddies Similarity \cite{BBS}.
Fig. \ref{fig:reranking} shows an example of the re-ranking method. 

\textbf{Re-Voting}. Ranked (or positional) voting system are suitable for Nearest Neighbor classification, where we can rank all our candidates. Re-Voting leverages the multiple short segments covering each query video, where each segment ranks all gallery IDs based on their distance from the query. During testing, each segment votes for the matching gallery IDs. The re-voting algorithm counts these votes by considering not only the first choices of each segment, but also lower-ranked matches.

Given a query video split into $M$ segments, let $\{s_i\}_{i=1}^M$ be the video segments. 
Let $g_j$ be a gallery ID, and let $r_{i,j}$ be the rank of ID $g_j$ for segment $s_i$. E.g. when $r_{i,j}=1$ it means that ID $g_j$ is ranked 1 for segment $s_i$
 
Of the several ranked voting methods we tested, the "Dowdall System" performed best: The re-voting score $V(g_j)$ is computed for all IDs $g_j$ using all $M$ segments in the video as follows:

\begin{equation}
    V(g_j) = \sum_{i=1}^{M} \frac{1}{r_{i,j}}
\label{eq:re-voting}
\end{equation}

For any segment the ID $g_j$ is at rank 1 it will add 1 to the vote for this ID, any time it is at rank 2 it will add 0.5 to the vote, etc. After re-voting for all IDs is completed, the ID that gets the highest vote is ranked 1, the ID than gets the second highest vote is ranked 2, etc.

\subsection{Can Latent Vectors Replace Skeletons?}
\label{sec:latent}

In \cite{Nguyen_2024_WACV} it was suggested to use a 3D mesh model of the human body, but instead of the actual mesh data the ReID used the latent values of the network generating the 3D mesh. Based on this experience, we tried to use the latent values of the skeleton models, rather than the skeletons themselves.

We extracted latent features from a pre-trained pose estimation model, specifically the hidden layers in BlazePose \cite{bazarevsky2020blazeposeondevicerealtimebody} and OpenPose \cite{cao2017realtime}. OpenPose yielded better results. We extracted activations after the \textbf{feat\_concat} layer, which merges intermediate feature maps across scales, capturing a rich pose representation. For each frame, we get a latent vector of size 864, and for a video segment of 50 frames we get a tensor $\vec{x} \in \mathbb{R}^{864 \times 50}$. A global average pooling across all frames in the segment gave the final embedding vector $\vec{z} \in \mathbb{R}^{864}$. Classification was done for each segment based on Nearest Neighbors in the Gallery. We then used the Re-Ranking and Re-Voting as described in Sec.~\ref{sec:rerank}. Although the results are reasonable, they are less effective than training the GCN encoder on the skeleton itself.

\subsection{Implementation details}

Pose estimation is performed using BlazePose \cite{bazarevsky2020blazeposeondevicerealtimebody}, which provides 33 human body joints per skeleton. The authors created a new topology by combining the super-sets from BlazeFace \cite{bazarevsky2019blazeface}, BlazePalm \cite{zhang2020mediapipe}, and Coco \cite{lin2014microsoft}. We have also successfully used OpenPose \cite{cao2017realtime} which provides 25 joints per skeleton, with about the same results.

The graph encoder $E$ described in Sec.~\ref{sec:GCN} is simplified compared to the original AAGCN \cite{shi2019two} as we removed the attention matrices. The pooling as described in Sec.~\ref{sec:GCN} is also different than the original pooling.


\textbf{Training and Testing.} The model was trained for 50 epochs using the Stochastic Gradient Descent (SGD) optimizer \cite{ruder2016overview} with a Nesterov momentum \cite{botev2017nesterov}. The learning rate was initialized to $1 \times 10^{-2}$ and reduced by a factor of 0.1 every 10 epochs. Training was conducted end-to-end on a single NVIDIA GeForce RTX 3070 GPU with 16GB RAM, taking approximately 1 hour. The implementation was done in PyTorch \cite{paszke2019pytorch}.

\section{Experiments}
\label{sec:experiments}

\subsection{Datasets and Evaluation Protocols}

We used the public Video-Based Clothes-Changing dataset CCVID \cite{gu2022clothes} to evaluate our method. CCVID contains 2,856 tracklets with 226 identities, each with 2 to 5 different suits. No distractors are present in CCVID. The training set includes 968 tracklets of 75 identities, while 834 tracklets are used as the query set, and the remaining 1,074 tracklets form the gallery set.

\textbf{Evaluation protocols.} Rank-k accuracy and mean average precision (mAP) are used to evaluate the performance of our method. We compute testing accuracy in two settings: (1) Clothes-Changing (CC), which uses only query samples that have matching gallery samples with different clothes. Additionally, gallery samples with the same identity and the same clothes are discarded; (2) Standard, which uses all query and gallery samples to calculate accuracy.

\subsection{Results}

\begin{table}[htb]
    \centering
    \begin{tabular}{|l|cc|cc|}
        \hline
        \multirow{2}{*}{Method} & \multicolumn{2}{c|}{Clothes-Changing} & \multicolumn{2}{c|}{Standard} \\
        \cline{2-5}
        & R-1 & mAP & R-1 & mAP \\
        \hline
        \hline
        InsightFace \cite{Deng_2020_CVPR} & 73.5 & - & \textbf{95.3} & - \\
        CAL \cite{gu2022clothes} & 81.5 & 79.6 & 82.6 & 81.3 \\
        ReFace \cite{arkushin2022reface} & 90.5 & - & 89.2 & - \\
        DCR-ReID \cite{10036012} & 83.6 & 81.4 & 84.7 & 82.7 \\
        SEMI \cite{Nguyen_2024_WACV} & 82.5 & 81.9 & 83.1 & 81.8 \\
        \hline
        OpenPose Latent & 76.2 & 80.4 & 70.2 & 78.0 \\
        \textbf{Ours} & \textbf{92.7} & \textbf{94.7} & 92.7 & \textbf{94.7} \\
        \hline
    \end{tabular}
    \caption{Comparison of quantitative ReID accuracy results on CCVID. Note the superior performance of our method on both the clothes-changing and the standard settings (also including same-clothes cases). Our results are identical in both cases as our method is the only one that is completely blind to appearance.}
    \label{tab:ccvid-results}
\end{table}

Quantitative results on the CCVID dataset \cite{gu2022clothes} are presented in Table \ref{tab:ccvid-results}, comparing our method with InsightFace \cite{Deng_2020_CVPR}, CAL \cite{gu2022clothes}, ReFace \cite{arkushin2022reface}, DCR-ReID \cite{10036012}, and SEMI \cite{Nguyen_2024_WACV}. Our approach, which uses only skeleton data, achieves superior performance across both Clothes-Changing and Standard settings (also including same-clothes cases). Our method outperforms all other approaches in the Clothes-Changing scenario, setting a new state-of-the-art. While InsightFace \cite{Deng_2020_CVPR} shows the best results in the Standard scenario, its performance drop in the Clothes-Changing setup indicates that changes in appearance can significantly impact its effectiveness. In contrast, our method's reliance on skeletal data ensures stable and superior performance, unaffected by appearance variations.


\subsection{Ablation Study}

\noindent\textbf{Spatio-Temporal Pooling}

As discussed in Sec.~\ref{sec:proposed-method}, we found that the Global-Average-Pooling (GAP) used in the original AAGCN paper \cite{shi2019two} is not optimal for ReID. We explored multiple pooling approaches. 
We evaluated off-the-shelf methods like GEM \cite{8382272}, SAG \cite{knyazev2019understanding}, and Global Average/Max Pooling, and more. In the MLP approach, we used two non-linear fully connected layers to reduce dimensions to four latent components. For the Weighted TopK method, we selected the top 1 frame temporally and the top 4 nodes spatially. Empirically, we found that best pooling was temporal pooling through stride in the temporal convolutions, and $L_3$ pooling (defined in Eq.~\ref{eq:lp-pool}) on all nodes in final layer. 

~\\
\noindent\textbf{Re-Ranking and Re-Voting}

\begin{table}[htb]
    \centering
    \begin{tabular}{|l|c|c|}
        \hline
        Method & R-1 & mAP \\
        \hline
        \hline
        GCN Only & 54.8 & 30.2 \\
        GCN with RR & 63.1 & 53.6 \\
        GCN with RV & 88.1 & 91.8 \\
        GCN with RR \& RV & 92.7 & 94.7 \\
        \hline
    \end{tabular}
    \caption{The impact of the Re-Ranking (RR) and Re-Voting (RV) components on the accuracy of ReID on the CCVID dataset.}
    \label{tab:ablation:rr-rv}
\end{table}

Table \ref{tab:ablation:rr-rv} evaluates the contributions of Re-Ranking (RR) and Re-Voting (RV) described in Sec.~\ref{sec:rerank}. Re-Voting has the largest impact on accuracy, indicating the importance of using all segments in a video clip. Still, adding Re-Ranking gives substantial improvement.

~\\
\noindent\textbf{Re-Voting}. 

The voting scheme in Sec.~\ref{sec:rerank} Eq.~\ref{eq:re-voting} gave to the $i$'th NN weight of $1/i$, and is called the "Dowdall System". We have also examined another common ranked voting scheme called "Borda Count". In this scheme the closest K-NN are examined, and the $i$'th NN gets a weight of $K-i$, and weight of $0$ for $i \ge K$. Results vary as a function of $K$, but never improved the results of the "Dowdall System" obtained in Eq.~\ref{eq:re-voting}. Also, in Eq.~\ref{eq:re-voting} there is no need to guess what is the optimal $K$. 
~\\

\noindent\textbf{Using Latent Features}

Table~\ref{tab:ablation:latent-skeleton} shows comparison of results using the latent features from the OpenPose model as described in Sec.~\ref{sec:latent}, to the our GCN descriptor derived from the actual skeleton. The GCN performs better. The results in this table follow the proposal in Sec.~\ref{sec:latent} to average all 50 descriptors of the 50 skeletons in a video segment. Alternative pooling methods of the 50 descriptors were tried, including the GRU method proposed in  \cite{Nguyen_2024_WACV}, but none performed better than the average pooling. 


\begin{table}[htb]
    \centering
    \begin{tabular}{|l|c|c||c|c|}
        \hline
        \multirow{2}{*}{Method} & \multicolumn{2}{c|}{OpenPose  Latent} & \multicolumn{2}{c|}{GCN} \\
        \cline{2-5}
        & R-1 & mAP & R-1 & mAP \\
        \hline
        \hline
        NN only & 48.0 & 12.0 & 54.7 & 30.2 \\
        NN \& RR \& RV & 76.2 & 80.4 & 92.7 & 94.7 \\
        \hline
    \end{tabular}
    \caption{Comparison of results using latent features derived from pretrained OpenPose to our proposed GCN features. First raw compares the classification for a single query segment using nearest neighbors, and the second row adds re-ranking and re-voting using all video segments.}
    \label{tab:ablation:latent-skeleton}
\end{table}

\section{Conclusion}

We have introduced an approach for Clothes-Changing ReID using only the joints in the human skeleton to compute distinctive ID signatures. In addition to Nearest Neighbor ranking, classical methods such as Re-Ranking and Re-Voting were used to achieved state-of-the-art results on the CCVID dataset. Despite the exclusion of appearance information, our results demonstrate the potential of body joints as a robust feature for ReID, even under the challenge of Clothing-Changes. The reduced use of appearance data also enhances privacy. 

{\small
\bibliographystyle{ieee_fullname}
\bibliography{egbib}
}


\end{document}